\documentclass[letterpaper, 10 pt, conference]{ieeeconf}  % Comment this line out if you need a4paper

\IEEEoverridecommandlockouts                              % This command is only needed if 
\usepackage{graphics} % for pdf, bitmapped graphics files
\usepackage{epsfig} % for postscript graphics files
\usepackage{mathptmx} % assumes new font selection scheme installed
\usepackage{times} % assumes new font selection scheme installed
\usepackage{amsmath} % assumes amsmath package installed
\usepackage{amssymb}  % assumes amsmath package installed
\usepackage{siunitx} % for SI units
\usepackage{textcomp}
\usepackage{gensymb} % for degree command
\usepackage[colorlinks=true]{hyperref}
\usepackage{booktabs}
\usepackage{multirow}
\usepackage{xcolor}
\usepackage{booktabs}
\let\labelindent\relax
\usepackage{enumitem}
\usepackage{xspace}
\usepackage{booktabs}
\usepackage{soul}
\usepackage[noadjust]{cite}
\usepackage{makecell}
\usepackage{multicol}
\usepackage{rotating}
\usepackage[percent]{overpic}

\usepackage{microtype}
\usepackage{contour}
\usepackage{courier}
\usepackage{algpseudocode}
\usepackage{algorithm}
\usepackage[subtle]{savetrees}

\usepackage{pifont}
\newcommand{\cmark}{\ding{51}}%
\newcommand{\xmark}{\ding{55}}%
\newcommand{\mypara}[1]{\vspace{2 mm}\noindent\textbf{{#1}}}

\title{\LARGE \bf Bag All You Need: Learning a Generalizable Bagging Strategy \\ for Heterogeneous Objects}

\author{Arpit Bahety$^*$$^1$ \quad Shreeya Jain$^*$$^1$ \quad Huy Ha$^1$ \quad Nathalie Hager$^1$ \quad \\  Benjamin Burchfiel$^2$ \quad  Eric Cousineau$^2$ \quad  Siyuan Feng$^2$ \quad  Shuran Song$^1$ \\
\href{https://bag-all-you-need.cs.columbia.edu/}{bag-all-you-need.cs.columbia.edu}
\thanks{$^*$ indicates equal contribution}
\thanks{$^{1}$ Columbia University $^{2}$ Toyota Research Institute }%
}

\begin{document}

\maketitle
\thispagestyle{empty}
\pagestyle{empty}

\begin{abstract}
We introduce a practical robotics solution for the task of heterogeneous bagging, requiring the placement of multiple rigid and deformable objects into a deformable bag.
This is a difficult task as it features complex interactions between multiple highly deformable objects under limited observability. 
%requires a  understanding of the objects geometry, material as well their interaction with the bag in order to  strategically plan the bagging actions. 
To tackle these challenges, we propose a robotic system consisting of two learned policies: 
a rearrangement policy that learns to place multiple rigid objects and fold deformable objects in order to achieve desirable pre-bagging conditions, and a lifting policy to infer suitable grasp points for bi-manual bag lifting. 
We evaluate these learned policies on a real-world three-arm robot platform that achieves a 70\% heterogeneous bagging success rate with novel objects. To facilitate future research and comparison, we also develop a novel heterogeneous bagging simulation benchmark that will be made publicly available. 

\end{abstract}

\section{Introduction}
\label{sec:introduction}
%Bag packing, while an effortless task for humans, is highly challenging for robots due to complex interactions between numerous rigid and deformable objects as well as severe self-occlusion and partial observability. \shuran{I feel the first sentence is repeating the next example, and repeating with the bullet points.} 
Imagine packing a bag for a picnic; we might first put several rigid objects (such as an apple and a water bottle) into the bag, fold deformable objects (such as a picnic mat and a T-shirt) and then place them on top of the bag opening. We must then lift the bag (another deformable object) in a way that these objects fall inside without spilling.
Successful completion of this task requires both a comprehensive understanding of the objects' physical properties and the capability to plan and integrate multiple manipulation skills. For instance, the robot's actions must take into account:

\begin{itemize}%[leftmargin=3mm]
    \item \textbf{Object geometry:} objects must be placed and oriented to fit into the bag opening. 
    
    \item \textbf{Object material:} large deformable objects, such as blankets, must be folded or crumpled into a compact configuration prior to packing. This requires manipulation strategies that are conditioned on object material (\textit{i.e.} rigid and deformable). %Note that a successful pre-bagging configuration does not always require all the objects to be fully inside the bag.  %Many deformable objects have large surface area in some configurations and must be carefully shaped to fit within the bag opening. Successful packing of these items often relies on folding or crumpling them into a smaller shape before packing. To accomplish this, a robot must employ manipulation strategies that condition on object material (i.e. rigid and deformable) to achieve a desirable ``pre-bag'' configuration. 
    
    \item \textbf{Inter-object dynamics:} the ultimate success of this task is determined jointly by object configurations and the robot's grasp on the bag during lifting. Crucially, when objects are partially inside a bag (for example, a mat on top of the bag opening), different lifting positions will result in different outcomes. Therefore, a successful approach must decide when a desired \textit{pre-bagging condition} is achieved and, if so, determine a good grasp location(s) to lift up the bag. Here, pre-bagging condition refers to when all objects are sufficiently inside the bag opening, and will fall into the bag with a proper lift. 
    % while taking into account the mutual configurations of each object and the bag itself. 
\end{itemize}

\begin{figure}[t]
    \centering
    \includegraphics[width=0.98\linewidth]{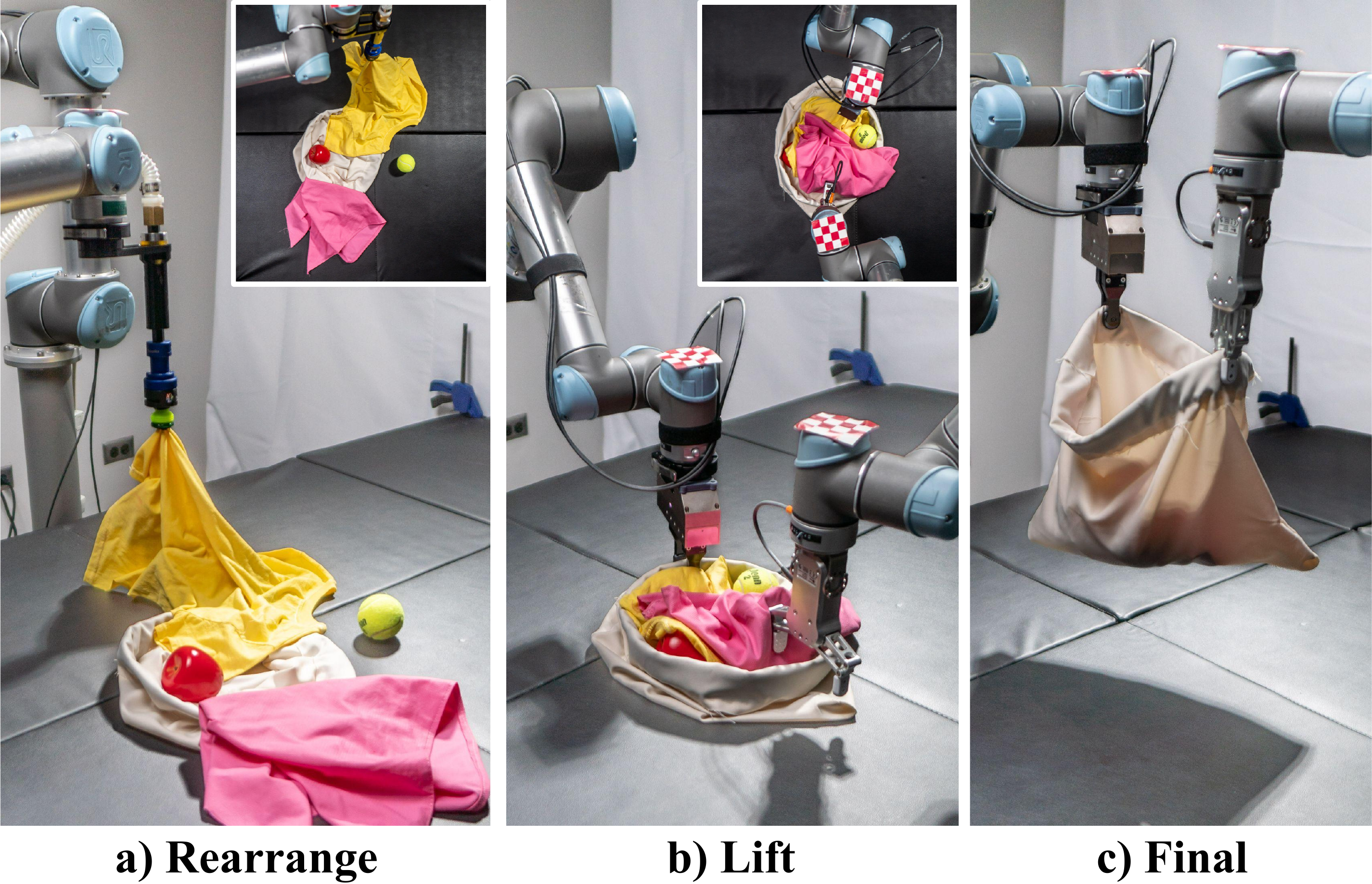}
    \vspace{-2mm}
    \caption{The \textbf{Heterogeneous Bagging Task} requires packing multiple rigid (e.g., the apple) and deformable objects (e.g., the T-shirt) into a deformable bag. The system must learn to (a) strategically manipulate these objects to achieve a feasible pre-bagging configuration. It also needs to (b) infer suitable grasp points from which to lift up the bag such that (c) the objects fall inside the bag.} 
    \vspace{-5mm}
    %https://docs.google.com/drawings/d/18KJCsIE08z6mNlNYXl-7MNhTnkalCvvobZvDqH6uoqc/edit?usp=sharing 
    \label{fig:teaser}
\end{figure}

\begin{figure*}[t]
    \centering
    \includegraphics[width=0.96\textwidth]{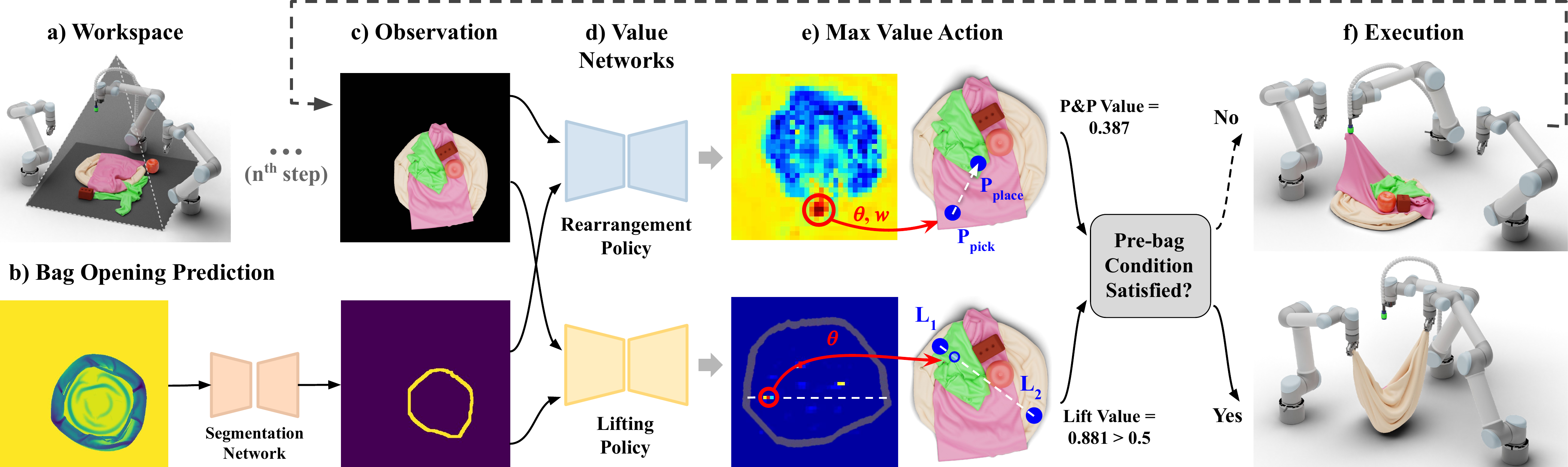}
    \vspace*{-1mm}
    \caption{\textbf{Method Overview.} Our system consists of (a) three robot arms and a top camera with a view of the workspace. A top-down depth image of the bag (before placing any other objects) is used to (b) predict the bag opening boundary. For each step, (c) a top-down RGB image and the predicted bag opening mask are input to (d) the rearrangement and lifting policies, which individually output (e) dense value maps and the action corresponding to the highest pick-and-place and lift score. If the pre-bagging condition is satisfied, the bag is (f) lifted from the lift points predicted by the lifting network. Otherwise, we (f) execute a rearrangement action and return to (c) for the next step.}
    \label{fig:method} \vspace*{-3mm}
    %  https://docs.google.com/drawings/d/1VwQYcLg_gNHPDR7_bfUEMWQt0-UHNRxR8FbbgcJRkZM/edit
\end{figure*}

Due to these difficulties, prior work focused either on only the lifting step of the process~\cite{seita2021initial} or considered a simplified scenario of packing only rigid items~\cite{zhao2021online, zhao2022learning}. 

%Solution: 
We seek to address these limitations and propose a system that tackles the complete bagging process for a diverse set of rigid and deformable objects --- a task we refer to as \textbf{heterogeneous bagging}. 
Our proposed approach consists of two learnable policies:
a \textbf{rearrangement policy} that uses sequential pick-and-place actions to rearrange or fold items (Fig.~\ref{fig:teaser}a) in order to achieve a suitable pre-bagging configuration and
a \textbf{lifting policy} that determines where to grasp and lift up the bag once pre-bagging conditions are met (Fig.~\ref{fig:teaser}b,c). We show that estimating the satisfaction of these pre-bagging conditions (required to decide when to stop rearranging and begin lifting) can be jointly performed by the two policies.
%performed at runtime by evaluating both learned policies jointly.

To accomplish this task on real hardware, we develop a representative simulation environment and use it to train both policies. Then, to facilitate a better bridge for the inevitable sim2real gap, we train a \textit{self-supervised} network that detects the bag opening from real-world depth images. These predictions are used as additional input to the rearrangement and lifting policies, allowing them to transfer more robustly from simulation, where they are trained, to the real world.  

We evaluate the learned policies with a real-world three-arm robot system with novel objects. The system is equipped with two types of end-effectors: a suction gripper, responsible for object rearrangement, on one arm and a parallel-jaw gripper, used to perform the bag lifting portion of the task, on the other two arms. We find that our proposed approach achieves a 70\% success rate for the heterogeneous bagging task.

The main contribution of this work is the development of the first real-world robot system for the task of heterogeneous bagging. To this end, we propose:  
%To the best of knowledge, we developed the first realworld robotics system for the task of heterogeneous bagging in unstructured environment. In support of that goal, we proposed:  
\begin{itemize}%[leftmargin=3mm]
    \item A self-supervised bag opening detection algorithm from depth images, whose pixel-wise supervision is automatically obtained through color images. This detection result enables robust sim2real transfer for downstream policies. 

    \item A learned rearrangement policy that strategically manipulates and reconfigures multiple rigid and deformable objects to satisfy required pre-bagging conditions. 
    
   \item A learned lifting policy that determines valid pre-bagging configurations and infers suitable grasp points for a bi-manual bag lifting action. 
   
   \item  A novel simulation environment and benchmark for heterogeneous bagging. The benchmark will be publicly available to facilitate future research and enable a fair comparison between heterogeneous bagging approaches. 
%\footnote{1
%Please visit \href{https://bag-all-you-need.cs.columbia.edu/}{bag-all-you-need.cs.columbia.edu} for experiment videos, code, simulation environment, and data.}  
\end{itemize}

\section{Related Work}
\label{sec:related_work}

\mypara{Rigid object packing.} 
Owing to numerous potential real-world applications, the problem of packing rigid objects has been extensively studied~\cite{zakka2020form2fit, zeng2020transporter}. %In order to formulate a feasible robot packing problem, several real-world constraints need to be addressed, such as collision-free paths, stability under gravity, and manipulation kinematics \cite{wang2019stable}.
In the offline setting, where the set of items and packing order are predetermined, prior works have primarily focused on exact algorithms~\cite{martello1998exact}, heuristics and metaheuristics~\cite{baker1980orthogonal, johnson1974worst}. In the online setting, where arbitrary items arrive sequentially and must be packed in the order they are received, deep reinforcement learning strategies~\cite{zhao2021online, zhao2022learning}, and the NDOP/QOP algorithm for the nondeterministic order setting~\cite{wang2020robot} have been used. However, all these approaches are limited to packing \textit{rigid}, generally cuboidal, objects into \textit{rigid} containers and are not suitable for deformable objects or non-rigid containers such as bags.
%packing rigid object into rigid box
%http://motion.cs.illinois.edu/papers/TASE2020-Wang-RobotPackingNondeterministic-preprint.pdf

\mypara{Cloth and rope manipulation.} % In contrast to rigid objects, deformable objects have high-dimensional configuration spaces and complex, non-linear dynamics. 
Early attempts at deformable object manipulation focused on methods for manipulating one-dimensional deformable objects such as ropes and cables~\cite{hopcroft1991case, saha2007manipulation, suzuki2021air,chi2022irp, nair2017combining, seita2021learning} and two-dimensional deformable objects such as fabrics~\cite{sanchez2018robotic, bersch2011bimanual, maitin2010cloth, osawa2007unfolding, cusumano2011bringing}.
%, mcconachie2020manipulating} - went to next page, add back if we're adding more references
% 
%While the unpredictable pose and physical properties of two-dimensional structures such as fabrics pose a challenge, they account for a majority of the research on deformable object manipulation due to extensive industrial, textile, and domestic applications. Some notable works involve bi-manual grasps \cite{sanchez2018robotic, bersch2011bimanual}, vision-based point detection and iterative grasping of corners \cite{maitin2010cloth}, and gravity for iterative grasping of the lowest hanging point \cite{osawa2007unfolding, cusumano2011bringing}. 
Data-driven techniques such as Reinforcement Learning and Imitation Learning have also been developed for cloth smoothing~\cite{ganapathi2021learning, lin2022learning}, folding~\cite{ebert2018visual, matas2018sim, jangir2020dynamic}, and unfolding~\cite{ha2022flingbot, seita2020deep}.
While our approach is inspired by some of these prior works in fabric folding, we address a significantly harder task that involves 3D deformable objects such as bags and complex interactions between multiple deformable objects. 

\mypara{Bag manipulation.} 
The manipulation of 3D deformable objects, such as bags, is an under-studied research area in robotics due to the inherent complexity and difficulty of the task. Initial work involved calculating the deformation characteristics of an object and determining the minimum lifting force through iterative lifting~\cite{howard2000intelligent}. Recent relevant studies involve grasping randomly or at maximum width to lift a bag using a physical robot~\cite{seita2021initial} or opening a deformable bag and maintaining the opened state using air-based blowing actions~\cite{xu2022dextairity}.

The most relevant work to our task is perhaps Seita et al.~\cite{seita2021learning}, where the task is to insert a rigid object into a deformable bag. But their approach is limited to handling a single rigid object placement and further simplifies the bag lifting task by attaching rigid beads around the bag opening. 
In contrast, our system can manipulate multiple objects (either rigid or deformable) and infer lift points for a fully-deformable bag directly from real-world RGB-D images, resulting in a more practical solution for real-world applications.

%In contrast to prior approaches that insert rigid objects into rigid containers \cite{zhao2021online, zhao2022learning, wang2020robot} or rigid objects into deformable bags \cite{seita2021learning}, our goal is to pack both rigid and deformable objects into a deformable bag. Moreover, we learn the grasp points and lift a fully-deformable bag, as opposed to random/maximum-width grasp points \cite{seita2021initial} or manipulation of bags through rigid beads at the opening \cite{seita2021learning}.

\section{Method}
\label{sec:method}
\subsection{Task and System Setup}
\label{subsec:method_setup}
% describe the task
We formulate the bagging task as follows: First, a bag is placed on a flat surface with its mouth open and facing upward. From this configuration, the robot perceives the bag and infers the bag opening (Fig.~\ref{fig:method}b). Note that this predicted bag opening remains constant throughout the episode. We then position all the objects randomly across the workspace (Fig.~\ref{fig:method}a). The robot manipulates and iteratively rearranges these objects to obtain a desirable pre-bagging configuration, estimates a pair of bag-lifting grasp points, and attempts to lift the bag. Finally, a bag-shaking primitive is executed to help the objects either drop inside the bag or fall out. We define success as no objects touching the surface of the workspace at the end of this sequence (\textit{i.e.} all objects are inside the lifted bag). Refer to Fig.~\ref{fig:method} for an overview of the pipeline.

% robot camera setup 
\textbf{Simulation benchmark.} Our simulation environment is built on top of the PyFleX bindings to Nvidia FleX~\cite{ha2022flingbot, xu2022dextairity, lin2020softgym, li2018learning}. We provide the functionality to load robot end-effectors, arbitrary cloth meshes, and arbitrary rigid objects. The simulated objects include our custom bag, primitive rigid geometries, a subset of the YCB dataset~\cite{calli2015benchmarking, calli2017yale}), rectangular cloths, and the CLOTH3D dataset~\cite{bertiche2020cloth3d} (Fig.~\ref{fig:objects}). The internal stiffness is uniformly sampled from$[0.85\text{kg/s}^2, 0.95\text{kg/s}^2]$, bag dimensions from $[0.25\text{m}, 0.40\text{m}]$, and cloth dimensions from $[0.20\text{m}, 0.60\text{m}]$. The object HSV color is uniformly sampled from $[0.0, 1.0]$, $[0.0, 1.0]$, $[0.5, 1.0]$ and the observations are rendered using Blender 3D. To generate different initial configurations of the bag and cloths, they are randomly picked and dropped on the surface. We also ensure that the cloths are larger than the bag opening and the objects are not entirely inside the opening.
% The simulated training objects include a bag, primitive rigid objects, a subset of the YCB dataset~\cite{calli2015benchmarking, calli2017yale}), and rectangular cloths (Fig.~\ref{fig:objects}). {\color{orange}The size, stiffness and color of the bag are randomized}. The object colors are randomized and observations are rendered using Blender 3D. {\color{blue}Q:Should we provide sampling range or provide more information here?}
% where the object HSV color is sampled uniformly between $[0.0, 1.0]$, $[0.0, 1.0]$, and $[0.5, 1.0]$ respectively.

Our real-world setup consists of three UR5 robot arms, one with a suction gripper for object rearrangement and two with parallel-jaw grippers for bag lift. The suction gripper allows for targeted manipulation of only the top cloth layers, preventing accidental bag grasps during rearrangement. To perceive the workspace, the robots obtain a top-down RGB-D image from an Azure Kinect sensor.

\begin{figure}
    \centering
    \includegraphics[width=0.96\linewidth]{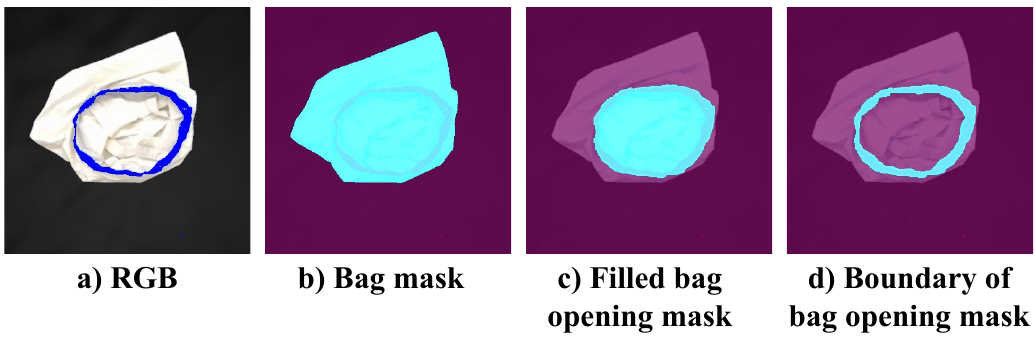}
    \vspace{-3mm}
    \caption{\textbf{Illustrating Different Masks Used.} This figure serves to clarify the terminology used for the different types of masks discussed in this paper.}
    % https://docs.google.com/drawings/d/1E1g8EudqivTmoL_97c266yelHAxCh1s_n5C0tsWYgyM/edit
    \label{fig:bag_masks} \vspace{-3mm}
\end{figure}

% maybe briefly talk about the bag segmentation and opening segmentation algorithm here

\subsection{Self-supervised Bag Opening Detection}
\label{subsec:opening_detection}
A reliable bag opening detection is critical to the success of our approach. 
% However, due to significant visual differences between the real-world and the simulated bag, an algorithm trained in simulation would face difficulty generalizing to real-world images. Meanwhile, pixel-wise bag opening annotations in the real world are expensive to obtain.
However, an algorithm trained in simulation would face difficulty generalizing to real-world images due to significant visual differences between the real-world and the simulated bag. Meanwhile, pixel-wise bag opening annotations in the real world are expensive to obtain.

\begin{figure}
    \centering
    \includegraphics[width=0.96\linewidth]{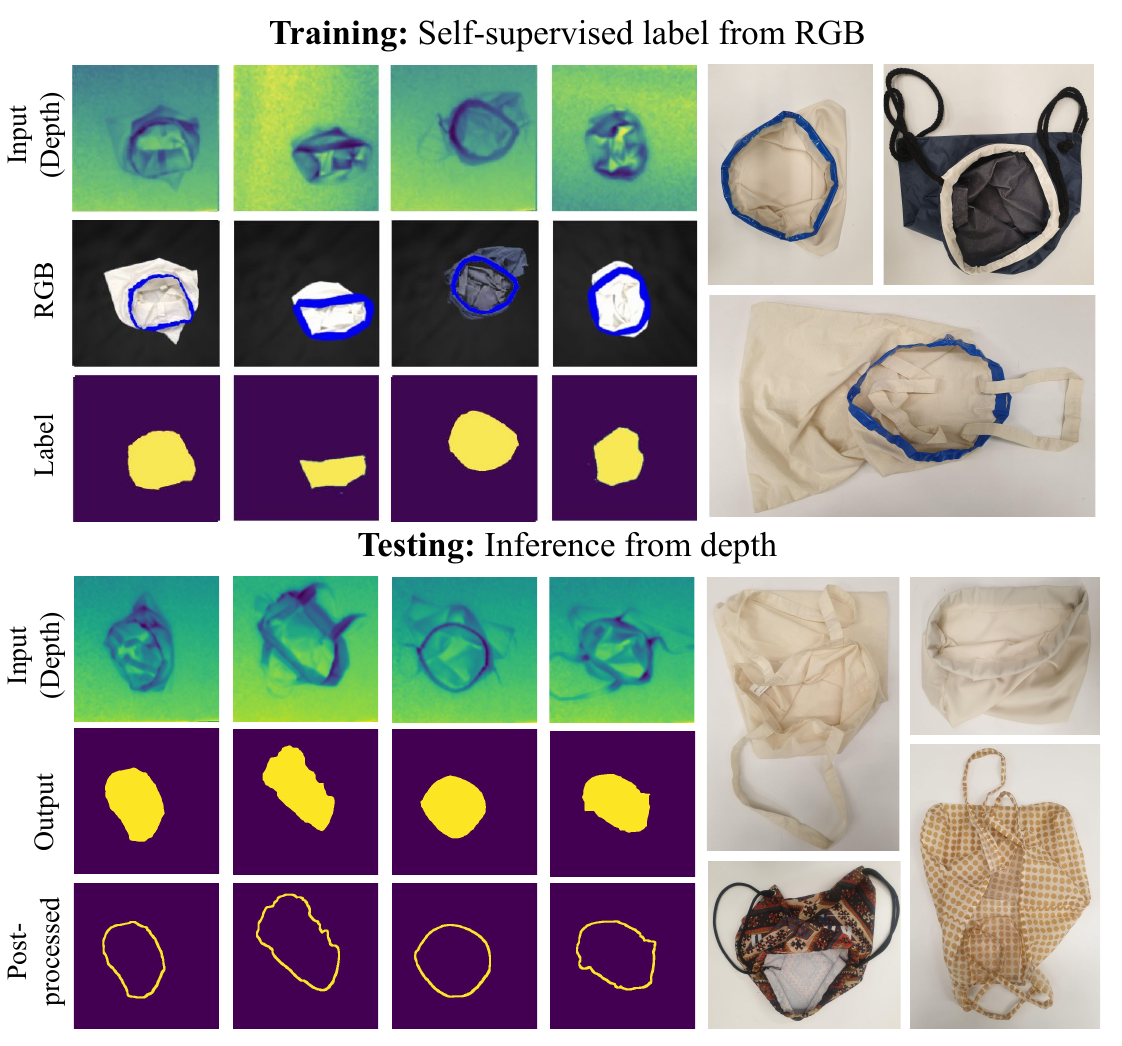} \vspace{-3mm}
    \caption{\textbf{Self-supervised Bag Opening Detection.}  We color the boundary of the bag opening (only during training) and use this color information to automatically obtain a pixel-wise label. At test time, the network inputs a depth image and outputs the predicted filled bag opening mask which is post-processed to obtain the boundary of the bag opening mask. We obtain 95\% intersection over union (IoU) for the test bags showing the robustness of our method. Note that the test bags do not have a colored opening.}
    % New Link: https://docs.google.com/drawings/d/1-vZGqv5tMsiKedJ_7Se-v0Zflq2cHuPt8V56MAO4FA0/edit
    \label{fig:bag_op_detection} \vspace{-3mm}
\end{figure}

% we propose a self-supervised model where the pixel-wise label is automatically obtained from RGB images of a bag with a colored opening. Concretely, as depicted in Fig.~\ref{fig:bag_op_detection}, we color the opening of the training bag in the real world blue and use this color information to obtain a segmentation label.
To tackle the above challenges, we propose a self-supervised model that predicts the bag opening solely using depth information, while training the model with pixel-wise labels obtained from RGB images of bags with colored opening. This is illustrated in Fig.~\ref{fig:bag_op_detection}, where we color the opening of the training bag in the real world and use this color information to automatically obtain the segmentation labels.  We use a filled bag opening mask (refer to \textit{(c)} in Fig.~\ref{fig:bag_masks} and \textit{Label} in Fig.~\ref{fig:bag_op_detection}) as supervision to avoid the class imbalance issue.
We then train a U-Net~\cite{olaf2015unet} based segmentation network with Dice Loss~\cite{Sudre2017GeneralisedDO} to predict the filled bag opening mask from depth images. 
% The network consists of four downsample layers followed by four upsample layers with skip connections 
During inference (Fig.~\ref{fig:bag_op_detection}), the segmentation network takes a depth image as input and outputs the predicted filled bag opening mask. The system subsequently post-processes this prediction to obtain only the boundary of the bag opening mask. The predicted filled bag opening mask and the post-processed boundary of the bag opening mask are used as additional inputs to the rearrangement and the lifting policy, respectively.
We utilize the intersection over union (IoU) metric over the filled bag opening mask to validate the performance of our bag opening detection model. The results indicate a 98\% IoU for training bags (200 images for 3 bags each) and 95\% IoU for test bags (25 images for 4 bags each), demonstrating that our model generalizes to novel and unseen test bags.

% In simulation, we can use the initial bag configuration to obtain the ground truth opening mask, which is then input to the rearrangement and lifting policies. This shared bag opening representation allows downstream policies to perform direct sim2real transfer without any real-world finetuning.  
In simulation, we can use the initial bag configuration to obtain the ground truth opening masks, which are then input to the rearrangement and lifting policies. This shared bag opening representation allows downstream policies to perform direct sim2real transfer without any real-world finetuning.

%Unfortunately, this subtask is not straightforward to learn via direct supervision. One possible approach is to use the ground truth bag opening information available in simulation to train a supervised segmentation model to predict the bag opening from RGB images. This method is limited, however, due to the fidelity of cloth-based simulation, which significantly lags behind that of rigid simulation, and results in a large sim2real gap. 

% solution: self-supervision from color image
% use image to provide label 
% use depth as input

% \begin{table}[t]
% \caption{{\color{orange}Self-supervised Bag Opening Detection Quantitative Results}} \vspace{-3mm}
% \centering
% \begin{tabular}{ l|cccc} 
% \toprule
%  & IOU\\ 
% \midrule
% Training bags & 98\%\\
% \midrule
% Test bags & 95\%\\
% \bottomrule
% \end{tabular}
% \label{Tab:real_world_result}
% \vspace{-3mm}
% \end{table}

\subsection{Rearrangement Policy}
\label{subsec:rearr_policy}
%\shuran{What's the input and output for this policy?}
A crucial part of the bagging task is to learn how to rearrange objects such that a good pre-bagging configuration is achieved; the object rearrangement policy is responsible for this.
To achieve a valid pre-bagging configuration, the policy must learn how to place objects inside the bag opening while ensuring no object is neglected.  %For example, the policy needs to learn where to place a bottle so that it does not fall away after the place action or how to manipulate a cloth to lie inside inside the bag opening.
%The policy also needs to determine which object to rearrange at a given time step and no object should be neglected. For instance large cloths such as a towel take up a lot of workspace --- and may not even fit entirely inside the bag opening --- while a small cup is comparatively trivial to arrange. In circumstances such as these, the policy must learn to not repeatedly rearrange a single difficult item and instead choose other objects to place inside the bag opening. 
Moreover, the policy needs to learn when to stop rearranging to avoid deteriorating a sufficiently good pre-bagging configuration and/or needlessly increasing the number of rearrangement steps.

% \begin{enumerate}
%     \item Where to iteratively place all the objects such that at the end of an episode, we have a good pre-bagging configuration.  Another example - where the policy should place a cloth (implicitly conditioned on the pick point) such that maximum volume of the cloth is inside the bag opening.
%     \item  For example, in the best case scenario, certain larger cloths such as large dress, might only partially fit inside the the bag opening, but the configuration could still be a good pre-bag configuration. In such scenarios, the policy needs to learn to not repeatedly rearrange the cloth but choose other objects to place inside the bag opening.
%     \item The policy also needs to learn the order of placing objects inside the bag opening. For example, placing a rigid object on top of a cloth and then rearranging the cloth could lead to the rigid object falling away.
%     \item Finally this policy needs to learn when to stop rearranging because further actions could deteriorate the pre-bagging configuration.
% \end{enumerate}

%  and outputs pick and place pixels, $P_{\mathrm{pick}}$, $P_{\mathrm{place}} \in \mathbb{R}^2$, which can be depth-projected to get 3D locations.

\textbf{Action parameterization and network.} 
The rearrangement policy takes a top-down RGB image and the bag opening mask as input and outputs pick and place pixels, $P_{\mathrm{pick}}$, $P_{\mathrm{place}} \in \mathbb{R}^2$, which can be depth-projected to get 3D locations.

We use the spatial action map policy formulation~\cite{ha2022flingbot, wu2020spatial, zeng2020tossingbot, zeng2018learning} and parameterize pick-and-place with $P_{\mathrm{pick}}$ representing the maximum value pixel from the value network's output, $\theta \in \mathbb{R}$ for the planar rotation and $w \in \mathbb{R}$ for the distance of $P_{\mathrm{place}}$ from $P_{\mathrm{pick}}$. The top-down color image and bag opening mask are concatenated ($H \times W \times 4$) and used to generate a batch of rotated and scaled observations ($t \times H \times W \times 4$). The network takes this batch as input and outputs a corresponding batch of predicted dense value maps ($t \times H \times W$). The \textit{argmax} of the value maps gives us $P_{\mathrm{pick}}$, rotation $\theta$ and scale $w$ (Fig.~\ref{fig:method}e), which are then used to compute $P_{\mathrm{place}}$. The value of each pixel corresponds to the predicted reduction of volume of all the objects outside the bag opening and thus, we choose the pixel with the maximum value. 
In our experiments, we use 12 rotations in the range of $[-180\degree, 180\degree]$ and 8 scale factors in the range of $[1.00, 2.75]$ at $0.25$ intervals (giving $t = 96$). The value network is implemented as a U-Net.

\textbf{Supervision and training.} The policy is trained via a self-supervised epsilon-greedy exploration in simulation. During a training episode, a random task (refer to Section~\ref{subsec:exp_setup}) is sampled and each rearrangement step is automatically labeled with the relative change in the volume of all the objects outside the bag opening: $\Delta v/v = (vol_{pre} - vol_{post})$ $/$ $\text{max}\{vol_{pre}, vol_{post}\}$, where $vol_{pre}$ and $vol_{post}$ are the pre-action and post-action volumes. A reward of $-0.5$ is given for pick points on the bag to ensure that $P_{\mathrm{pick}}$ is not on the bag during inference.
Note that this supervision signal (\textit{i.e.} object particles) is neither available nor required during real-world deployment. The robot manipulates objects with the rearrangement policy until no valid grasp is predicted (\textit{i.e.} the highest value action is not on any object) or reaches the maximum interaction iterations (\textit{i.e.} 10). The network is supervised via MSE loss between predicted and actual $\Delta v/v$ and is trained using the Adam optimizer~\cite{kingma2015adam} with a learning rate of 1e-3 and a weight decay of 1e-6. The network is trained to convergence, which takes around 40K simulation steps or about 32 hours on 4 NVIDIA RTX3090s. 

\subsection{Lifting Policy}
% \textbf{Goal of this policy: }
% \begin{enumerate}
%     \item Determining a good pre-bag configuration and thus, deciding when to stop rearranging
%     \item Learning where to grasp to lift up the bag
% \end{enumerate}
The final step of a bagging task is to lift the bag such that all objects fall inside. Accordingly, the lifting policy must predict a pair of points that are likely to be graspable and enable a successful lift. Secondly, it must determine when to take over from the rearrangement policy because a valid pre-bagging configuration has been achieved. Oftentimes, there is a significant nuance to this determination; for instance, a valid pre-bagging configuration may have a T-shirt’s sleeve extend beyond the bag opening but have its center of mass inside.
% Not all pairs of points on the bag opening would lead to success and so, the policy must predict a pair of lift points that are likely to be graspable and enable a successful lift.

\textbf{Action parameterization and network.} 
 % A lift action is parameterized by two points $l_1, l_2 \in R^3$, which denote the two lift points for a dual-arm robot system. The z dimension is inferred from depth information. 
The lifting policy takes a top-down RGB image and the bag opening mask as input and outputs a lifting score with the corresponding lift points, $L_1$, $L_2 \in \mathbb{R}^2$, which can be depth-projected to get 3D locations.
 To constrain the system to predict lift points on the estimated mouth of the bag, we formulate the action parameterization as $<P_{\mathrm{lift}}, \theta>$, where $P_{\mathrm{lift}}$ is a pixel that lies on a line and $\theta \in \mathbb{R}$ determines the slope of the line. The intersection of the line with the bag opening gives the two lift points, $L_1$, $L_2$ (Fig.~\ref{fig:method}e). To minimize collisions, we ensure that the distance between $L_1$ and $L_2$ does not exceed a given physical limit and is greater than the safe distance. 
 
The concatenation of the top-down color image and bag opening mask is used to produce a batch of rotated observations ($t \times H\times W \times 4$). This is input to the network and a corresponding batch of dense value maps ($t \times  H \times  W$) is predicted. Each pixel in each value map contains the value of the action parameterized by that pixel’s location, providing $P_{\mathrm{lift}}$, and the rotation applied to the input observation, providing $\theta$. The pixel value corresponds to the predicted probability of lifting success. At runtime, the robot selects the action with the highest predicted value. 
In our implementation, we use 12 rotations in the range of $\theta$ to $[-90\degree, 90\degree]$ (giving $t=12$). The value network is implemented as a U-Net.

\textbf{Supervision and training.} 
% The lifting policy is trained via self-supervised epsilon-greedy exploration in simulation where each lifting step is labelled as a 1 for a success and 0 for a failure. The tasks for training the network are generated from our trained rearrangement policy. The network is supervised via a binary cross-entropy loss and is trained using the Adam optimizer with a learning rate of 1e-3 and a weight decay of 1e-6. The policy is trained for around 100K simulation steps which takes around 72 hours on 4 NVIDIA RTX3090s.
The tasks for training the lifting network are generated from our trained rearrangement policy. Each lifting step is labeled as a 1 for success and 0 for failure, where success is defined as no object touching the surface of the workspace after a successful lift. The policy is trained similarly to the rearrangement policy but supervised via a binary cross-entropy loss. The training takes around 100K simulation steps or about 72 hours on 4 NVIDIA RTX3090s

% \subsection{Verifying Pre-bagging Conditions}
\subsection{Determining Lifting Conditions}
% \label{sec:stop}

% Even when not all objects are completely inside the bag opening, they can fall into the bag with a proper lift.
% In these cases, our pre-bagging condition enables more efficient bagging by allowing early-stopping rearrangement.
% Pre-bagging can be verified in two cases, 1) when the lift score is higher than a lift score threshold, or 2) when the policy predicts a termination action.
% We use a lift score threshold of 0.9 in simulation and 0.5 in real world to account for the lift-score network's lower confidence in out-of-distribution inputs.

The robot uses the output of both policies to determine when desirable pre-bagging configurations are achieved, thereby allowing early-stopping rearrangement. This is beneficial since each policy is trained with different objectives: there are cases when the rearrangement policy predicts an action to further reduce the volume of objects outside the bag opening while the lifting policy predicts that the current configuration is sufficient for success. In this scenario, it is inefficient and unnecessary to continue rearranging items. 
At other times, the lifting policy may not be certain that the pre-bagging conditions are met, but the rearrangement policy infers that there is no action that can improve the configuration. In this case, the system should proceed with the lifting action. 

At each step, the lifting score is evaluated to achieve this fused pre-bagging determination. If the lifting score is greater than a threshold, the rearrangement stops and the lift action is executed (Fig.~\ref{fig:method}). Otherwise, the rearrangement policy predicts an action, and if the action is to terminate, then the bag is lifted from the points corresponding to the highest score predicted previously by the lifting policy.
We use a lift score threshold of 0.95 in simulation and 0.5 in real world to account for the lift-score network's lower confidence in out-of-distribution inputs.

\section{Evaluation}
\label{sec:evaluation}
\subsection{Experiment Setup}
\label{subsec:exp_setup}
We use three metrics to evaluate system performance:
1) Success rate (\textbf{SR}): \textit{successful\_episode} / \textit{total\_episodes}, where a successful episode is defined in Section~\ref{subsec:method_setup}.
2) Average fraction of objects inside the bag (\textbf{AvgF}): The number of objects not touching the ground after lifting and shaking the bag divided by the total number of objects in that episode, averaged across all episodes.
3)  Average episode length (\textbf{AvgL}): The total number of rearrangement steps plus the lifting step in an episode, averaged across all episodes. We evaluate on the following task setups:
two (1r1c), three (1r2c, 2r1c), four (1r3c, 2r2c, 3r1c), and five (2r3c, 3r2c) objects, where \#r and \#c denote the number of rigid objects and cloths.

\begin{figure}[t]
    \centering
    \includegraphics[width= 0.97\linewidth]{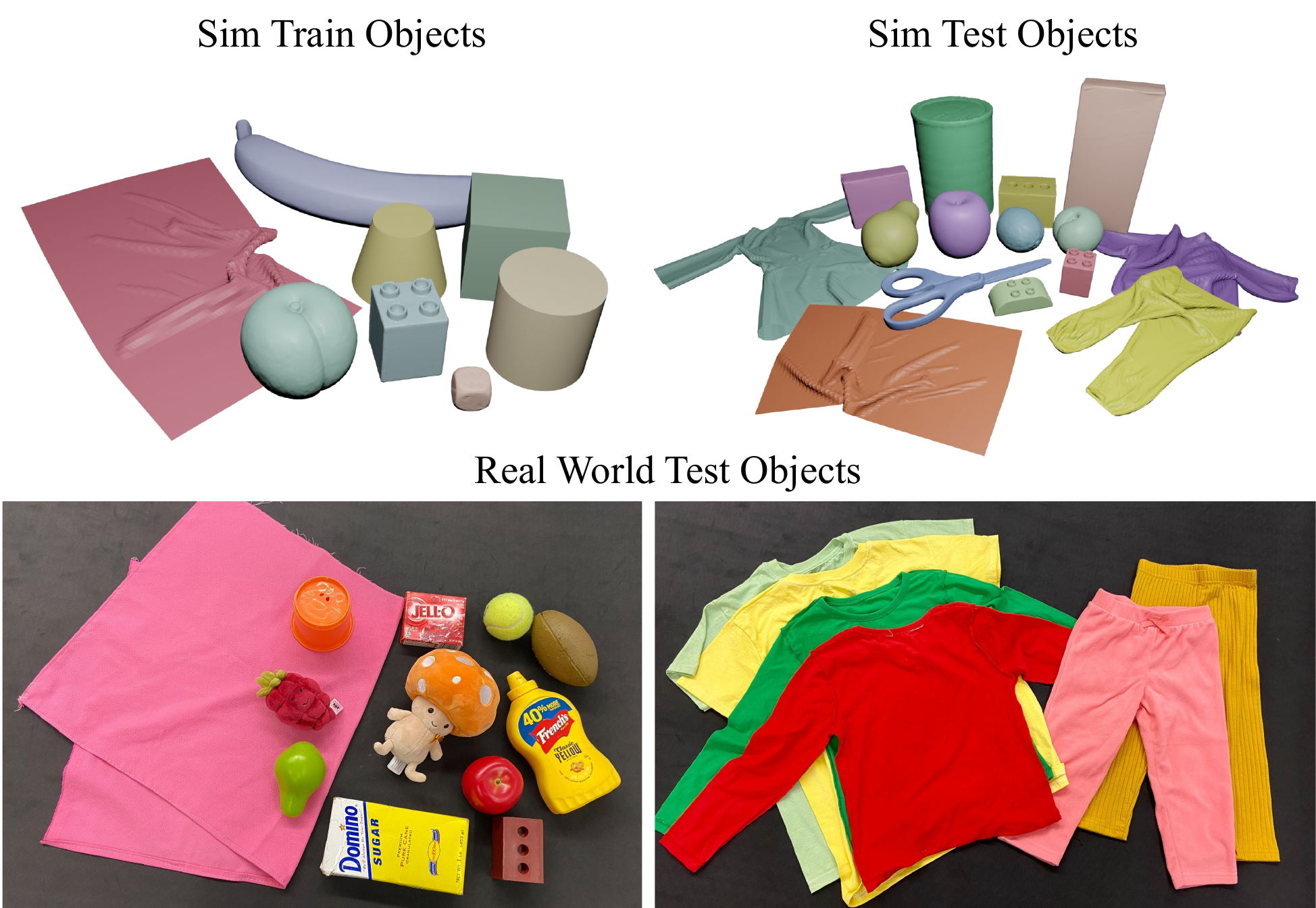}
    \vspace{-2mm}
    \caption{\textbf{Train and Test Objects} used in our experiments. The policy is always tested with novel objects not seen during training.}  
    \label{fig:objects}
    % https://docs.google.com/drawings/d/1erdw2mjOV0C2r4QRedCndxz_GRf72ySttX1gqo9SRBk/edit
\end{figure}

\textbf{Simulation tasks.} 
During testing, we use a subset of YCB objects (different from training) and CLOTH3D~\cite{bertiche2020cloth3d} (Fig.~\ref{fig:objects}). Each task has a bag that is kept open on the ground and 2-5 objects arranged in an arbitrary configuration in the workspace. Refer to Section \ref{subsec:method_setup} for details about task generation. 

\textbf{Real-world tasks.} 
The testing objects are a subset of YCB, soft toys, T-shirts, pants, and rectangular cloths (Fig.~\ref{fig:objects}). We run 5 different task configurations for the 2 objects case and 10 configurations for each of the 3-5 objects cases.

\textbf{Baseline algorithms.}
Since there are no existing methods for the task of heterogeneous object bagging, we design strong heuristic baselines for comparison.

\begin{figure}[t]
    \centering
    \includegraphics[width= 0.97\linewidth]{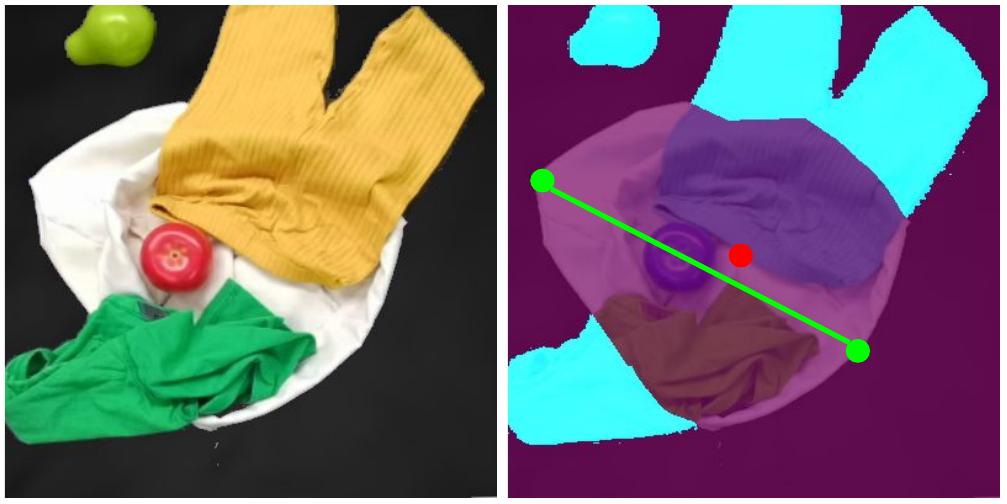}
    \vspace{-2mm}
    \caption{\textbf{Heuristic Rearrangement and Lifting Strategies in Real-world.} The rearrangement pick point is chosen from the part of object mask that is non-overlapping with the bag mask (blue region) and the place point is the centre of the bag mask (red circle). The lift points are obtained from the maximum-width lifting strategy~\cite{seita2021initial} (green circles)}  
    \label{fig:heuristic}
    % https://docs.google.com/drawings/d/1zNEGCZqv7XL7kgOR1-nZhS_rYTJ2j-JcN5DQLRszKGM/edit
\end{figure}

\begin{itemize}
\item Heuristic rearrangement policy: 
% We want a strong heuristic which picks objects outside of the bag opening and places them at a point inside the opening.
In both simulation and real-world, we design a strong rearrangement heuristic that picks objects from outside the bag opening and places them inside in a way that most likely reduces the volume of objects outside the bag.
The heuristic selects a random point on the part of object mask that is non-overlapping with the filled bag opening mask and places it at the center of the bag opening.
In real-world experiments, where the ground truth filled bag opening mask is unavailable, we use the bag mask instead, as shown in Fig. \ref{fig:heuristic}.
The rearrangement is terminated and the bag is lifted when either no pick point is found or the episode length reaches 10.

% \item Heuristic rearrangement policy: In addition to the information that our policy receives, the heuristic policy receives object masks as input. As shown in Fig. \ref{fig:heuristic}, the policy selects a random point on the part of object mask that is non-overlapping with the bag mask (filled bag opening mask in simulation) and places it at the center of the bag mask (filled bag opening mask in simulation). 
% {\color{orange} This is a strong heuristic because placing objects at the center will most likely reduce the volume of objects outside the bag opening.}
% The heuristic policies in simulation differ from real-world due to the availability of ground truth bag opening.
%\st{Intuitively, this heuristic is reasonable since placing objects at the center would likely reduce the volume of objects outside the bag opening.}

\item Heuristic lifting policy: We develop a strong lifting heuristic in accordance with the information available in each domain. In simulation, with the availability of ground truth bag opening, the heuristic randomly chooses two points on the boundary of the bag opening, with the constraint that the distance between them is greater than a threshold. In real-world experiments, due to the unavailability of a ground truth bag opening, the heuristic uses the maximum-width lifting strategy proposed by Seita et al.~\cite{seita2021initial} (Fig. \ref{fig:heuristic}).  
Note that this heuristic policy does not determine when to stop rearranging.
%{\color{orange} The heuristic in simulation is stronger than that in the real world due to the availability of ground truth bag opening in simulation.} 
\end{itemize}

%%%%%%%%%%%%%%%%%%%%%%%%%%%
\subsection{Experimental Results}

\begin{figure*}[t]
    \centering
    \includegraphics[width=0.95\linewidth]{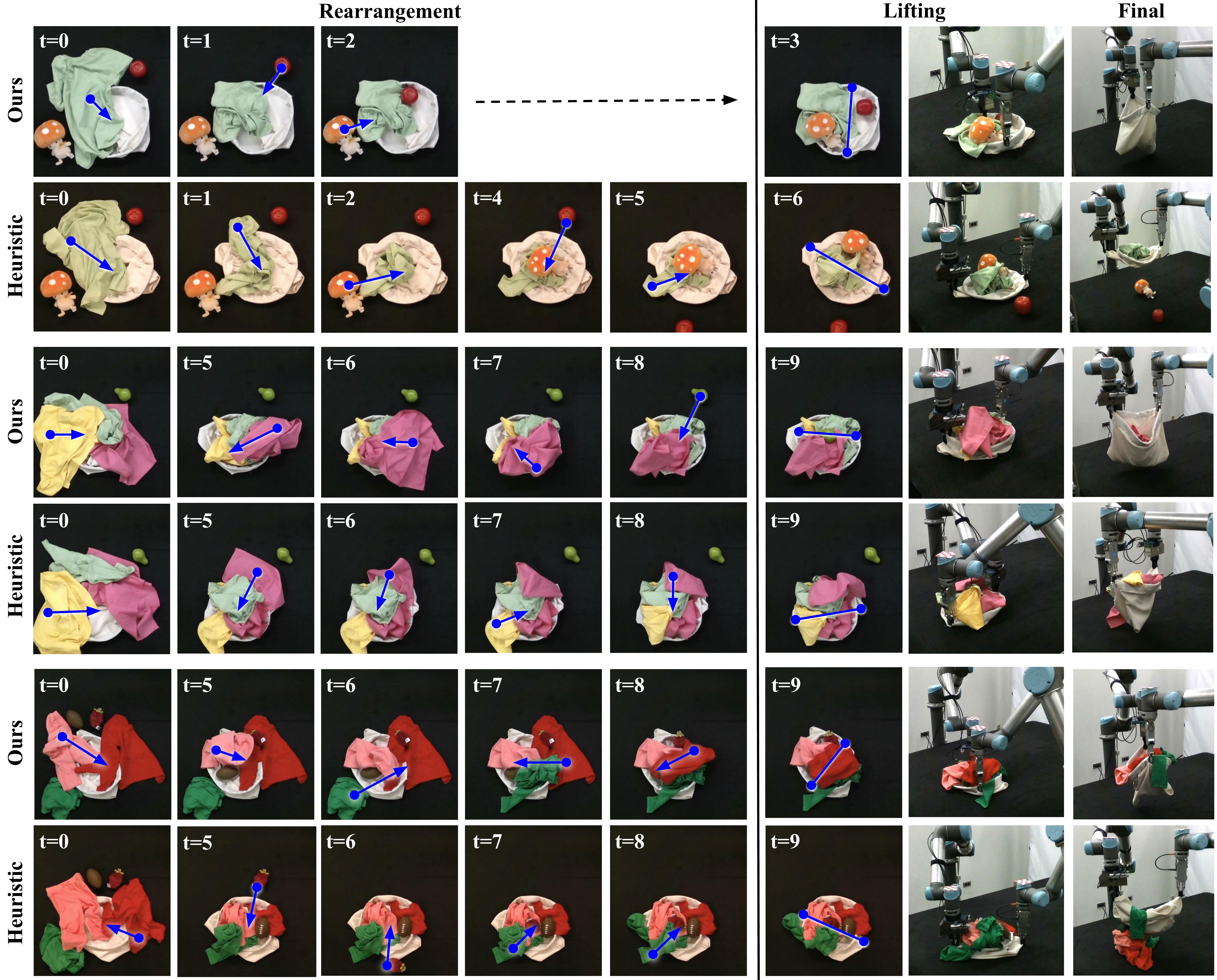}
    %https://docs.google.com/drawings/d/1fb3DJWly__sSLGlL4FfuSji5zQRGbfXMoIaCCg9XKAQ/edit
    \vspace{-3mm}
    \caption{\textbf{Real-world Bagging Results.} The first two examples illustrate how our policy is successfully able to rearrange all objects whereas the heuristic policy fails (the apple rolls away in the first example and in the second, the heuristic repeatedly rearranges the cloths and never the pear). In the third example, even with a good pre-bagging configuration, a heuristic lift fails as it does not find good lift points. More robot videos can be found on the   \href{https://bag-all-you-need.cs.columbia.edu/} {project website}.}
    \label{fig:results}
    \vspace{-3mm}
\end{figure*}

\textbf{Benefits of learned rearrangement policy.}
% describe H \& L  v.s. L \& L
Comparing the second and the last rows in Table~\ref{Tab:sim_result} shows the advantages of a learned rearrangement policy. We observe that as the number of objects increases, our method significantly outperforms the heuristic, signifying the importance of learning multi-object interactions for rearranging objects.
%describe why learned rearrangement is important 
%\todo{todo}
Real-world experiments also showcase the effectiveness of the learned rearrangement policy. As can be seen in Table~\ref{Tab:real_world_result}, our policy performs consistently better than the heuristic. Having a rearrangement policy that learns which objects to rearrange and where to place them helps to achieve a good pre-bagging configuration. We also find that learning when to stop rearranging to avoid unnecessary steps or prevent further actions from deteriorating the pre-bagging configuration helps improve the performance of the system.

\begin{table}[h]
    \begin{center}
        \addtolength{\tabcolsep}{-1pt}
        \caption{Success rate (\%) in simulation. }
        \begin{tabular}{ lll|cccc }
            \toprule

            \multicolumn{2}{c}{Learned}  & \multirow{2}{*}{Train Tasks} & \multicolumn{4}{c}{Number of Objects}                                                                 \\
            Rearr.                      & Lift                    &                                       & 2             & 3             & 4             & 5             \\
            \midrule
            \xmark                      & \xmark                  & -                                     & 75.0          & 50.5          & 23.3          & 11.0          \\
            \xmark                      & \cmark                  & mixed                                 & 89.0          & 73.5          & 45.3          & 28.5          \\
            \cmark                      & \xmark                  & mixed                                 & 80.0          & 61.0          & 35.3          & 22.5          \\
            \midrule
            \cmark                      & \cmark                  & single-rigid                          & 83.0          & 67.0          & 48.0          & 32.0          \\
            \cmark                      & \cmark                  & single-cloth                          & 89.0          & 76.0          & 55.0          & 39.0          \\
            \midrule
            \cmark                      & \cmark                  & mixed                                 & \textbf{92.0} & \textbf{81.5} & \textbf{75.6} & \textbf{61.5} \\
            \bottomrule
        \end{tabular}
        \label{Tab:sim_result}
    \end{center}
    \vspace{-6mm}
\end{table}
\vspace{-0.4mm}
\begin{table}[h]
    \begin{center}
        \caption{Sensitivity to bag opening in simulation. }
        \begin{tabular}{ l|cccc }
            \toprule
            \multirow{2}{*}{Average filled bag opening IoU} & \multicolumn{4}{c}{Number of Objects} \\
            
             & 2 & 3 & 4 & 5             \\
            \midrule
            100\% (ground-truth) & \textbf{92.0} & \textbf{81.5} & \textbf{75.6} & \textbf{61.5} \\
            95\% (= real-world test bag IoU) & 90.0 & 79.5 & 72.5 & 58.0 \\
            93\% ($<$ real-world test bag IoU) & 89.0 & 78.0 & 72.0 & 58.5 \\
            \bottomrule
        \end{tabular}
        \label{Tab:sim_bag_ablation}
    \end{center}
    \vspace{-3mm}
\end{table}

\textbf{Benefits of learned lifting policy.}
% describe L & H  v.s. L & L
Comparing the third and the last rows in Table~\ref{Tab:sim_result} illustrates the advantages of a learned lifting policy relative to the baseline (40\% improvement). Despite being strong due to the availability of ground truth bag opening, the heuristic performs worse than our learned policy because it is unable to learn the critical relationship between lift points and object configurations. It is also incapable of determining when a desirable pre-bagging configuration is achieved and rearrangement can be halted. 
Our real-world experiments corroborate the effectiveness of a learned lifting policy as well (Table~\ref{Tab:real_world_result}).
%describe why learned lifting policy is important 
%\todo{todo}
%Real-world experiments showcase the effectiveness of the learned lifting policy as well. Aside from the learned rearrangement policy, the learned lifting policy is the second contributor to the success of our policy in our experiments. The learned lifting policy infers the bag opening and then finds two optimal lift points on the bag opening. Without these predictions by the learned lifting policy, even a good pre-bagging configuration wouldn't be enough for a successful bagging. Furthermore, alike rearrangement policy, the lifting policy also helps to learn when to stop rearranging and lift by learning to differentiate a good pre-bagging configuration from a bad one. 

\textbf{Effects of different training tasks.}
% \todo{todo}
In order to understand the impact of training task selection, we conducted experiments with two other training scenarios - a single rigid object in each configuration and a single cloth in each configuration. As depicted in Table~\ref{Tab:sim_result}, the mixed training scenario performs the best, aligning with our intuition that training on a single object does not allow the policy to reason about multi-object interaction. Furthermore, the single-rigid policy performs worse than the single-cloth policy as it fails to learn appropriate folding strategies for cloths.
% Interestingly, the single cloth model outperforms the single-rigid model. We hypothesize that since rearranging deformable objects is a multi-step process whereas rearranging rigid objects is single-step, learning on deformable objects generalizes to the rigid case better than the reverse direction.

% describe when heuristic policy could work and when it cannot. Show both fold and lift need to learn and the effectiveness of it
% As can be seen in Table, heuristic rearranging policy works well with lesser number of objects as placing in centre 

% {\color{orange}\textbf{Effects of different training rewards.} In order to study the impact of the reward function we compare our particle-based reward to a vision-based reward.
% %In order to study the efficacy of a reward function based on ground truth state we compare our particle-based reward to a vision-based reward. 
% In particle-based reward, the volume of the objects are computed using the particles of objects available in the simulation whereas in vision-based reward, the volume is computed using the pixels of objects. The rest of the reward function remains the same (refer to \ref{subsec:rearr_policy}).}
% The last two rows in Table~\ref{Tab:sim_result} show that particle-based reward outperforms vision-based reward as computing volume using particles is more accurate that using pixels.

\textbf{Sensitivity to bag opening.} To examine the impact of accurate bag opening, we evaluated our method under two scenarios - a non-ablated version employing ground-truth bag opening as input, and an ablated version using a perturbed bag opening. The ablation technique involved randomly sampling bag opening vertices and transforming neighboring vertices using a Gaussian distribution centered on the selected points. Table \ref{Tab:sim_bag_ablation} indicates that using ground-truth bag opening only marginally improves the performance of our learned policy in comparison to a perturbed bag opening with 95\% filled bag opening IoU, equivalent to our real-world bag opening detection results from Section \ref{subsec:opening_detection}. Furthermore, additional perturbation of the bag opening to 93\% IoU has no substantial effect on the success rate.

\textbf{Real-world experiments.}
We directly evaluate our simulation-trained rearrangement and lifting policies with real-world robots. To promote policy transfer from simulation to reality, we perform background substitution that is consistent with the simulation environment. 
% For the heuristic baselines, we obtain the bag mask before any objects are present by removing the static background; we also use object masks to remove predicted lift points that overlap with objects.
We obtain the bag mask (for heuristic) and predict the bag opening (for ours) before any objects are placed. We also use object masks to remove predicted lift points that overlap with objects.
Table~\ref{Tab:real_world_result} shows the performance, averaged over all test episodes, for each task category. Our policy outperforms the heuristic by as much as 60\% on the \textbf{SR} metric. Moreover, the significantly higher \textbf{AvgF} metric (84\% vs 22\% in the 5 objects case) indicates that our policy drops fewer objects outside the bag.

Fig.~\ref{fig:results} illustrates examples of our system and the heuristic approaches operating in the real world. 
% \todo{Add explanation here for each example}
The first two rows demonstrate how heuristically placing all objects at the center of the bag can lead to failure. While our rearrangement policy learns to arrange each object relative to other objects inside the bag opening, the heuristic method places the apple on top of the soft toy, causing it to fall and roll outside the workspace (step 4, second row). This example also exhibits how our lifting policy learns that a partially protruding cloth can still be a valid pre-bagging configuration (green T-shirt). 
In the second example (the third and fourth rows), our policy successfully determines which object to rearrange so that a desired pre-bagging configuration is obtained. Despite the presence of three cloth objects partially outside the bag opening, our policy rearranges the pear (step 8, third row). In contrast, the heuristic keeps rearranging the cloth and overlooks the pear even though it is clearly outside the bag opening.
The third example shows that even when the heuristic rearrangement policy is able to achieve a good pre-bagging configuration, learning from where to lift the bag is crucial. Our policy is able to infer the bag opening and predict optimal lift points, which lead to success, whereas the lift locations chosen by the heuristic lead to failure.

\begin{table}[t]
\caption{Real-world Results.} \vspace{-3mm}
\centering
\begin{tabular}{ ll|cccc} 
\toprule
 & \multirow{2}{*}{Metric} & \multicolumn{4}{c}{Number of Objects} \\ 
 & & 2 & 3 & 4 & 5 \\
\midrule
Heuristic & SR $\uparrow$ & 60\% & 20\%& 20\%& 0\%\\
 & AvgF $\uparrow$  & 80.0\% & 43.3\%& 37.5\%& 22.0\% \\
 & AvgL $\downarrow$ & 3.8 & 5.6 & 5.9 & 8.5  \\
 
\midrule

Ours & SR $\uparrow$ & 80.0\% & 80.0\%& 70.0\%& 50.0\%\\
 & AvgF $\uparrow$  & 90.0\% & 93.3\%& 87.5\%& 84.0\% \\
 & AvgL $\downarrow$ & 3.6 & 4.5 & 5.7 & 8.2  \\
\bottomrule  
\end{tabular}
\label{Tab:real_world_result}
\vspace{-3mm}
\end{table}

\textbf{Failure modes and limitations.} 
The most common failure case was during bag grasping (accounting for 50\% of real-world failures). For instance, the robot would inadvertently grasp a cloth close to one of the predicted lift points and lift the cloth instead of the bag. Closing the loop between the rearrangement and the lifting policy to incentivize rearranging objects away from potential lift points might help resolve this issue.  
Another observed failure mode is suboptimal object rearrangement (accounting for 32\% of real-world failures). For instance, a rigid object placed upon a pile of cloth is generally unstable. This failure is caused by the single-step training reward, causing the rearrangement policy to not explicitly consider object stability for future interactions.

Our method also makes certain assumptions that could be relaxed in future works: It assumes that the objects are placed in the scene only after
the bag opening inference is performed. It also assumes that the bag is open and flat on a surface, potentially requiring an additional system~\cite{xu2022dextairity, autobag} to achieve this initial configuration. 
% It also assumes that the objects are placed in the scene only after the bag opening inference is performed which can be relaxed by training a bag opening detection model on a larger scale with bag and objects both in the scene.
%It is also limited to 2D pick and place; future work could consider a more expressive 6-DoF action space capable of inserting objects from the side into a partially open bag held up with one arm. 
%Such an action space may also help address grasping failures caused by object proximity to selected grasp points as noted above.

\section{Conclusion}
We propose and evaluate a real-world robotic system for the task of heterogeneous bagging which involves placing a diverse set of multiple rigid and deformable objects into a deformable bag. The system includes a self-supervised segmentation network trained to infer the bag opening, a rearrangement policy, and a lifting policy. We demonstrate the effectiveness of our method by comparing it against strong baselines on a set of real-world tasks. 
We hope that this work encourages others in the field to explore the challenging task of heterogeneous bagging and believe that the lessons learned in this setting have strong potential to transfer to other dexterous manipulation tasks involving the interaction of multiple highly deformable objects.
% \input{text/appendix}

%%%%%%%%%%%%%%%%%%%%%%%%%%%%%%%%%%%%%%%%%%%%%%%%%%%%%%%%%%%%%%%%%%%%%%%%%%%%%%%%



\section*{ACKNOWLEDGMENT}
This work was supported in part by the Toyota Research Institute, NSF Award \#2143601, \#2037101, and \#2132519. We would like to thank Google for the UR5 robot hardware. The views and conclusions contained herein are those of the authors and should not be interpreted as necessarily representing the official policies, either expressed or implied, of the sponsors.

%%%%%%%%%%%%%%%%%%%%%%%%%%%%%%%%%%%%%%%%%%%%%%%%%%%%%%%%%%%%%%%%%%%%%%%%%%%%%%%%

\bibliographystyle{IEEEtran}
\bibliography{references}

\end{document}